\documentclass[conference]{IEEEtran}
\IEEEoverridecommandlockouts
\usepackage{cite}
\usepackage{amsmath,amssymb,amsfonts}
\usepackage{algorithmic}

\usepackage{textcomp}
\usepackage{xcolor}
\usepackage[english]{babel}
\usepackage{amsmath}
\usepackage{amssymb}
\usepackage{gensymb}
\usepackage[final]{graphicx}
\usepackage{subfig}
\usepackage[utf8]{inputenc}
\usepackage{cite}
\usepackage{multicol}
\graphicspath{{./images/}}

\def\BibTeX{{\rm B\kern-.05em{\sc i\kern-.025em b}\kern-.08em
    T\kern-.1667em\lower.7ex\hbox{E}\kern-.125emX}}
\begin{document}

\title{Wall Detection Via IMU Data Classification In Autonomous Quadcopters}

\author{\IEEEauthorblockN{J. Hughes, D. Lyons}
\IEEEauthorblockA{\textit{Department of Computer and Info. Science} \\
\textit{Fordham University}\\
Bronx, New York, USA \\
jhughes50@fordham.edu}

}

\maketitle

\begin{abstract}
An autonomous drone flying near obstacles needs to be able to detect and avoid the obstacles or it will collide with them. In prior work, drones can detect and avoid walls  using data from camera, ultrasonic or laser sensors mounted either on the drone or in the environment. It is not always possible to instrument the environment, and sensors added to the drone consume payload and power - both of which are constrained for drones.

This paper studies how data mining classification techniques can be used to predict where an obstacle is in relation to the drone based only on monitoring  air-disturbance. We modeled the airflow of the rotors physically to deduce higher level features for classification. 
Data was collected from the drone's IMU while it was flying with a wall to its direct left, front and right, as well as with no walls present.
In total 18 higher level features were produced from the raw data. We used an 80\%, 20\% train-test scheme with the RandomForest (RF), K-Nearest Neighbor (KNN) and GradientBoosting (GB) classifiers. Our results show that with the RF classifier and with 90\% accuracy it can predict which direction a wall is in relation to the drone. 
\end{abstract}


\section{Introduction}
State of the art autonomous flying drones can fly to a preprogrammed destination while avoiding objects with no input from a user [13]. Such approaches leverage drone or environment-mounted camera, laser, or ultrasonic sensors to detect potential collisions. 
However, one critical barrier to more widespread drone usage in industry is battery life [13]. Because the drones must be light to fly they must have small batteries, and thus they have a short flight time. The previously mentioned sensors for object avoidance cut down on the drone’s battery life substantially. 
In our prior work we showed wind currents encountered by a drone can be detected using a classifier [1,2]. The wind induces greater pose disturbance, and this can be detected from the data gathered by the drone’s internal sensors. Additionally, the drone experiences greater pose disturbances when it is close to large objects like walls. This is because backwash from the wall interferes with the airflow created by the rotors of the quadcopter. This interference causes a similar effect to that of a wind current.

The idea behind this paper is that the backwash wind current created by a quadcopter being near an object interferes with the pose of the drone which can be used to detect which side of the drone the interfering object is located. This eliminates the need for the camera, laser or ultrasonic sensors
and thus increases the battery life. In a typical RC teleoperated drone scenario, the onboard computer would have to work no harder than now, since  the data points could be gathered and sent to the remote computer. The base computer would then do the calculation and prediction which can then be sent to the onboard driver program to trigger the drone to avoid an object.

As a step towards more general collision detection, this research address the interaction of the drone with perpendicular walls because they provide a good surface for 
air reflection from the backwash from the quadcopter’s rotors. Our initial focus on walls also reflects the fact that  it is likely that drones will encounter these when they fly autonomously in buildings. 
In this study, a quadcopter was flown perpendicular to the walls to collect clear data. 

The drone used in this research is the crazyflie 2.0, a lightweight drone ideal for indoor use and with a straightforward programming interface [7]. The quadcopter has an inertial measurement unit sensor (IMU). This measures the gyroscope in the $(x,y,z)$  planes, acceleration with an accelerometer again in the $(x,y,z)$ planes. From the gyroscope and accelerometer roll, pitch and yaw angles can be calculated for recording. There is also a barometer to measure air pressure. On the bottom of the quadcopter is a Flow Deck that measures the drone’s $z$ position optically. The drone also has an Loco Positioning System (LPS) Deck that measures its $(x,y,z)$ cartesian position with respect to base beacons in all eight corners of the flight area. While the Flow Deck and LPS are crucial for the data collection and data mining experiments, they are not used at all in the proposed collision detection.
 
We 
studied the drone's airflow 
in the vicinity of a wall using talcum powder and imaging to help us generate more meaningful features for classification. To gather the data the drone was flown along a wall to its left, right and front as well as with no wall present. After collecting the data we generated features based on the 
airflow study conducted earlier. Multiple tests were then run with the data. First, the presence of a wall was tested for, 
i.e., could the classifier detect a difference between the left and right data and the no wall data?
Next, the classifier was used to detect left versus right wall and left
 versus right versus front wall and finally left versus right versus front versus no wall. These tests were repeated 
 with
 three different classifiers: RandomForest, GradientBoosting and NearestNeighbor, and the results are presented here.

\section{Literature Review}

There have been prior studies done on the ground effect in quadcopters. A ground effect occurs when the drone is flying near the ground and the air that is being pushed down from the rotors hits the ground and backwashes up onto the drone, causing instability [17]. This effect has been studied extensively to make quadcopter landing and takeoff more stable.
Drone 
on
drone 
air disturbance
was studied in [2]. This project had one drone fly underneath the other and could successfully detect
the presence of the upper drone
using only the internal sensors of the bottom quadcopter.
The researchers in [1] were able to detect
when a drone encountered
 wind gusts using only the internal sensors of the drone. They had the drone fly along a path that encountered an air current from a fan and built a classifier to detect the air disturbance from the fan.

These two studies show that the internal sensors can be used to detect the presence of 
air disturbances. Specifically, [2] shows that drones can detect wind currents created by rotors. These papers differ from this project in the respect that they do not predict from which direction the wind is coming. This is pertinent to building a truly autonomous drone that does not rely on outside sensors. Its important to note that in our initial research we tried a purely bottom-up data mining approach, testing on only the raw data from the IMU. We found this did not work: that it was necessary to generate higher level, more descriptive features, and giving rise to the research presented in this paper.

Yoon et. al. and Diaz et. al. from NASA's Ames Research Center used Computational Fluid Dynamics (CFD) to model the airflow of a drone's rotors in [5,6]. They show that the velocity of the air underneath the rotor takes the form of a cylinder. The air moves faster within the cylinder in the shape of an hourglass. This is important because it is our hypothesis that when a drone is near a wall the air underneath it's rotors is getting pushed outward and then interacting with the wall, causing pose disturbances. Their work was done using the DJI Phantom 3 which is large and powerful. Thai et. al. [4] also did CFD for the DJI Phantom 3. Unlike the others, their work shows one instance of the drone's propeller spinning and shows that the velocity has a helical flow pattern.
They also show that the airflow from the rotors has 
an hourglass shape. 

Other researchers at McGill University have worked on wall detection without using additional sensors in [3]. In this study, drones with rotor guards fly directly at the wall. They investigate at what speed can the quadcopter travel, hit the wall and still recover.The idea behind this study is to have an autonomously flying drone that hits walls, recovers and moves in a different direction after the collision. This project wants to avoid the collision and just use the air disturbance from the wall to fly autonomously. Also, not all drones fly with rotors guards, and without them the drone would just crash. 

\section{Modeling}

Physical modeling of the airflow of the rotors was done to help better understand the effect the nearby wall had on the drone's stability and what phenomena could cause the instability. Greater knowledge of how the drone behaves close to the wall helped create higher level features that capture the instability.

\subsection{Physical Modeling}
 
In order to physically examine the airflow of the rotors, a drone was fixed to a spindle and placed against a black background.  The rotors were engaged at incremental thrust levels. Talcum powder was dropped from above into the drone's rotors. Images were taken of the resulting powder motion, and this was used to reveal airflow interaction with a vertical surface brought into proximity of the drone. 
\begin{figure}[htp]
	\centering
	{\includegraphics[width=0.4\textwidth]{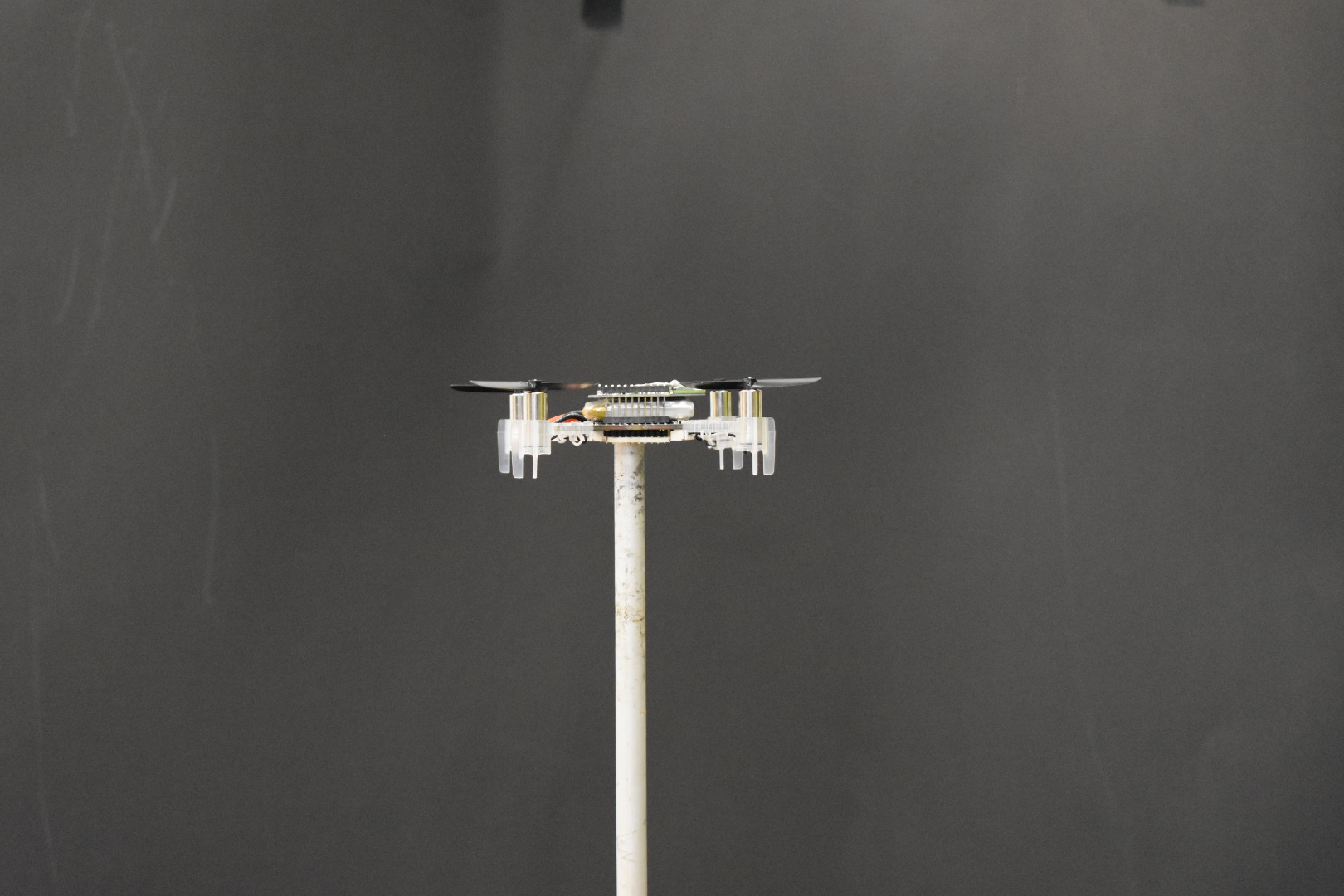}\label{fig:f1}}
	\caption{Experimental setup}
\end{figure}
The powder was dropped into the rotors from 0.5 meters above. The rotors were tested at thrust levels starting at 25\% and working up to 80\%. Figure 1 shows the experimental setup of the drone.
\begin{figure*}[h]
	\centering
	\subfloat[]{\includegraphics[width=0.3\textwidth]{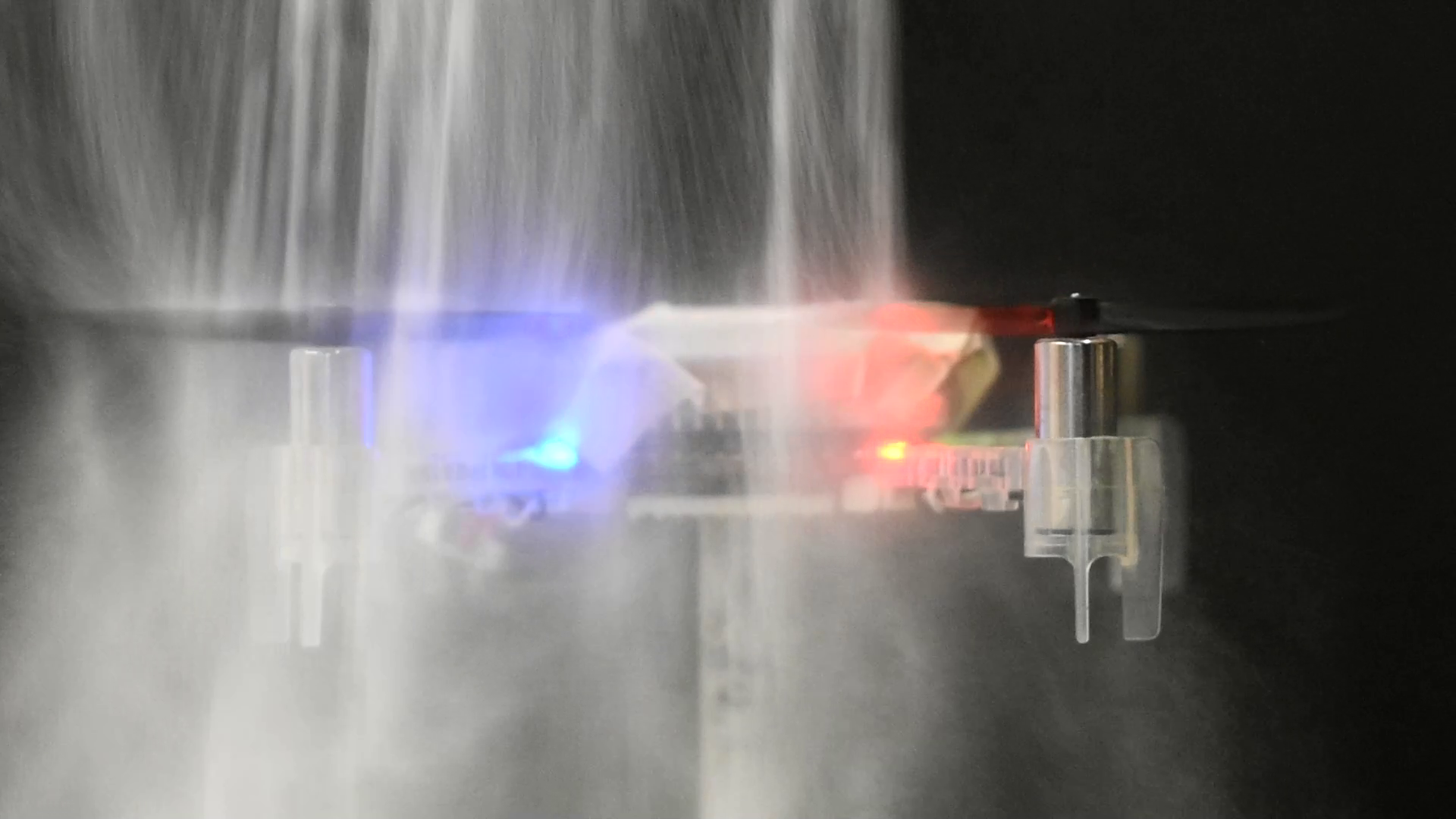}\label{fig:f1}}
	\hfill
	\subfloat[]{\includegraphics[width=0.3\textwidth]{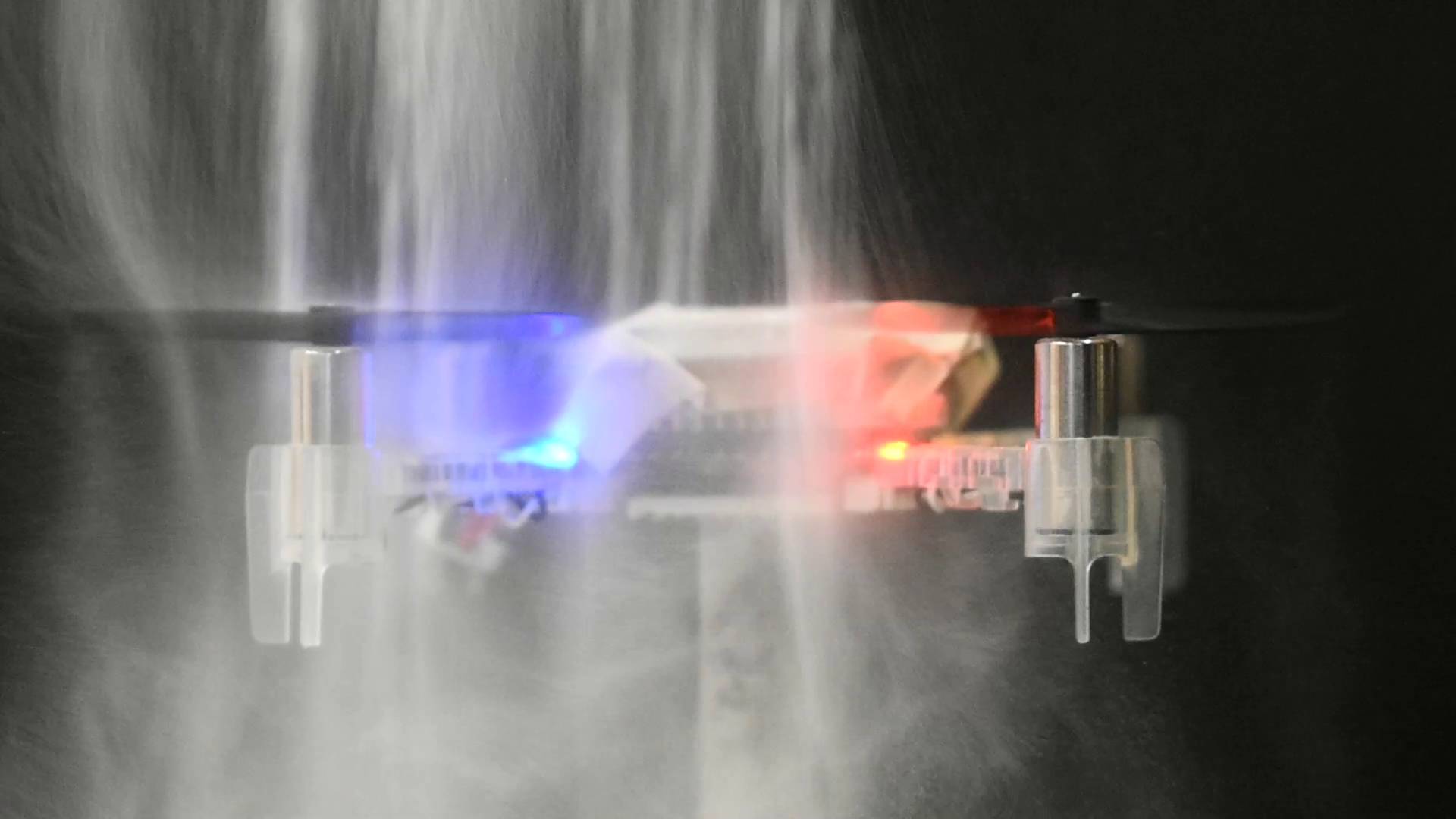}\label{fig:f2}}
	\hfill
	\subfloat[]{\includegraphics[width=0.3\textwidth]{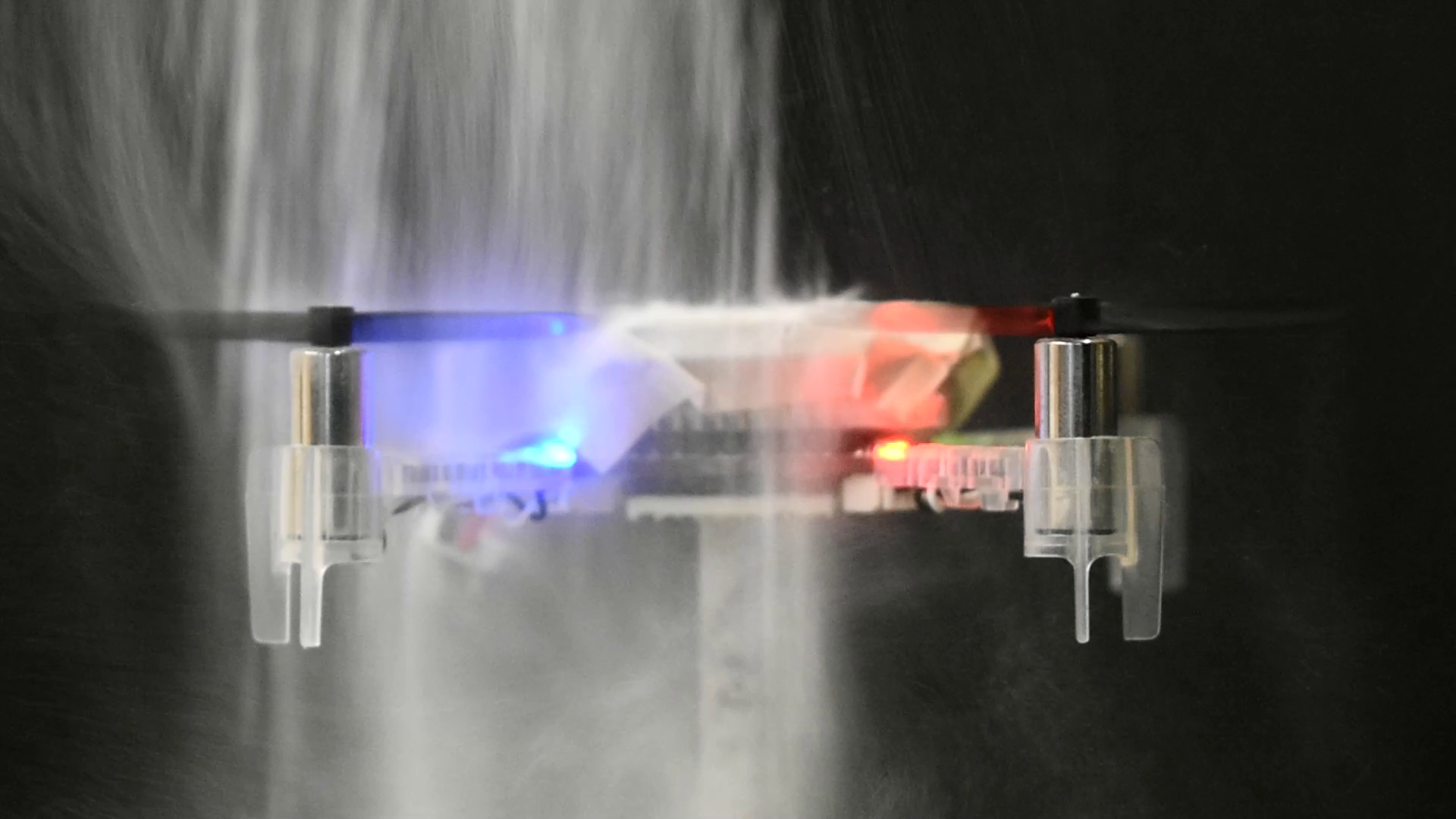}\label{fig:f3}}
	\caption{Sequence of frames from video showing airflow at 80\% thrust}
\end{figure*}  
\begin{figure*}[htp]
	\centering
	\subfloat[]{\includegraphics[width=0.3\textwidth]{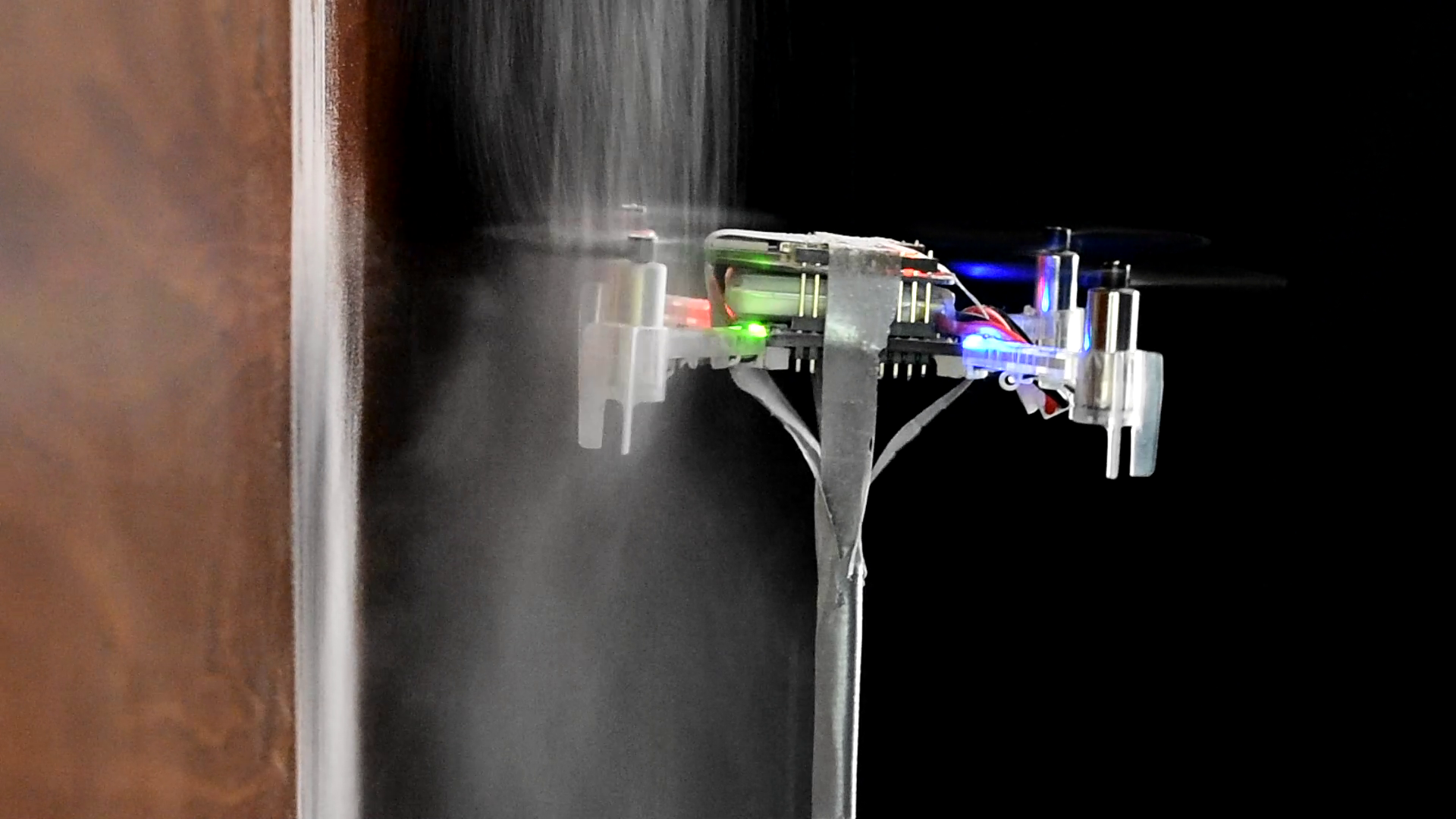}\label{fig:f1}}
	\hfill
	\subfloat[]{\includegraphics[width=0.3\textwidth]{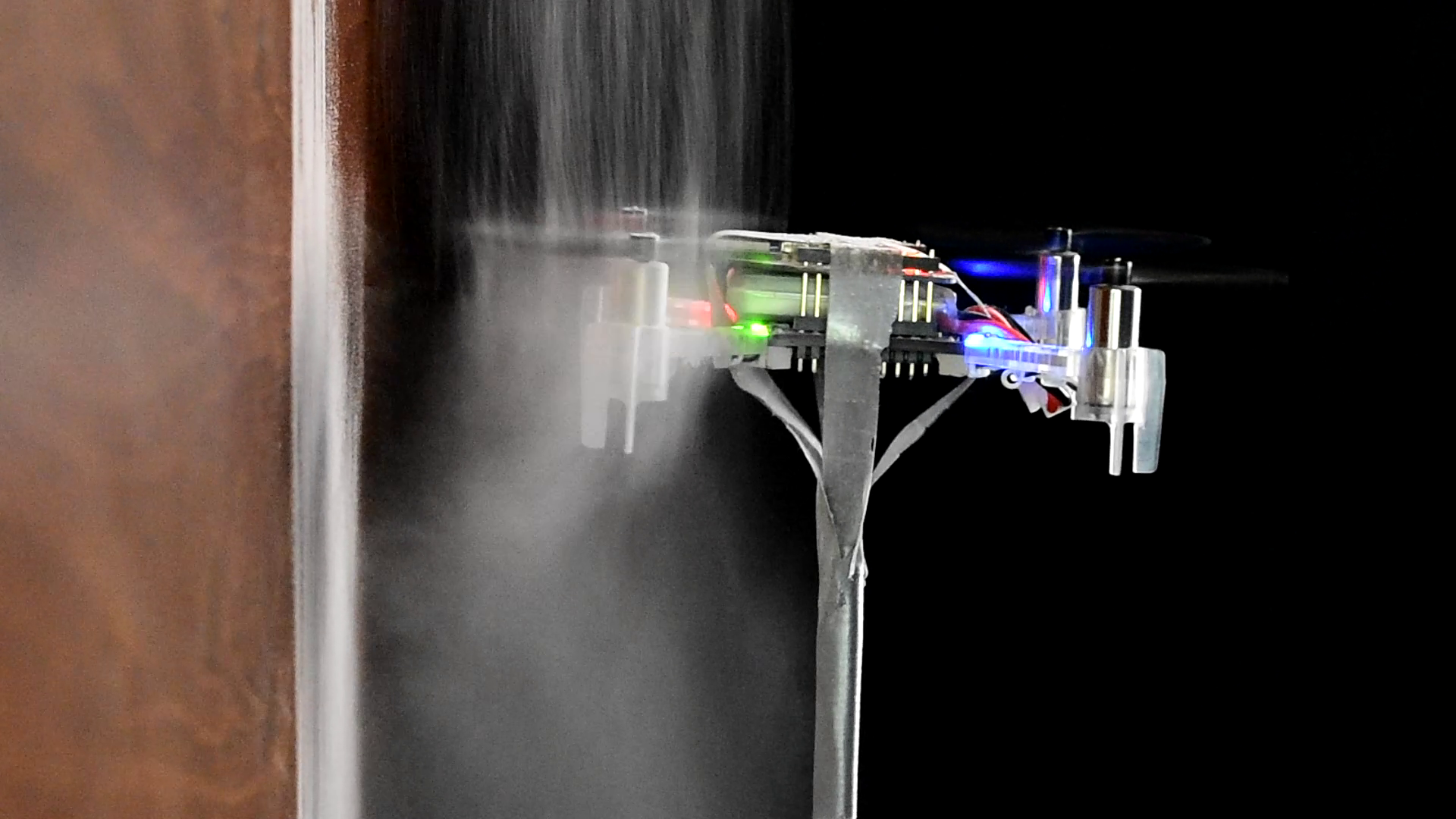}\label{fig:f2}}
	\hfill
	\subfloat[]{\includegraphics[width=0.3\textwidth]{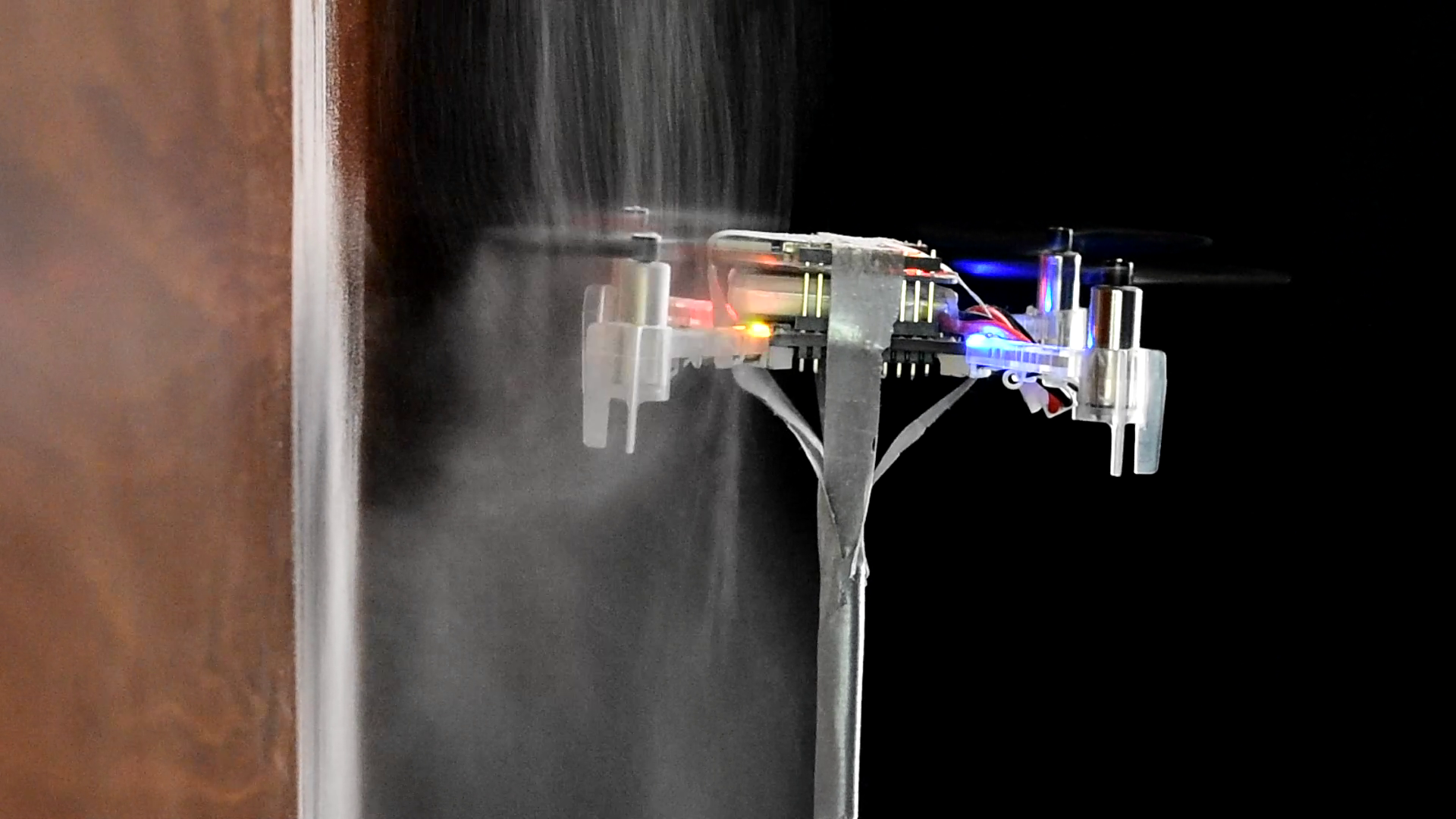}\label{fig:f3}}
	\caption{Sequence of frames with wall at 80\% thrust}
\end{figure*}
From the photos in Figures 2 and 3, we can see that particles of talcum powder form an area where there is a higher concentration of powder which follows an hourglass shape underneath the rotor shown best in figure 2c. The process was repeated but with a wall near the drone to examine how the airflow from the quadcopter interacts with the wall. The drone was placed two rotor diameter-lengths away from the wall, which is about 0.1m.

The hourglass shape that we observed is confirmed by the CFD in [6], which shows areas of higher pressure forming the same shape we observed. More specifically, they show that area of higher pressure is on the "surface" of the hourglass shape rather than filling a volume. This supports our hypothesis that the air is being pushed downward and outward toward the wall creating a backwash with the wall. In Figure 3, we can see the hypothesized backwash effect from the wall indicated by the more dispersed powder further underneath the wall. The larger cloud of powder can interrupt the air flowing downward, and we propose that this causes an 'artificial' ground effect. The lift is compromised on one side of the drone, resulting in larger than nominal pose disturbances. We hypothesize that it should be possible to tell which side of the drone this occurs on by monitoring higher-level features of the platform pose angles.

\subsection{Feature Generation From Modeling}

From the modeling it was deduced that the gyroscope would be the most effected IMU sensor on the drone. Indeed, when we look at the data we can see that the pose disturbance is shown more in the gyroscope than in the accelerometer.

We hypothesize that the unsteadiness of the drone occurs because of an artificial ground-effect. It is well studied that when drones are near the ground they become unsteady because the air from their rotors hits the ground and backwashes towards the quadcopter. In our case the air is hitting the wall and then dispersing, as air does when it hits an object. This dispersion causes a mass of air underneath the drone acting as the "artificial ground". The proposition is shown in Figure 4. Since the "ground-effect" is only on one side of the drone there is likely a difference in gyroscope, accelerometer and  stabilizer reading between a left wall, right wall and front wall. This is why we deduce that the gyroscope would be the IMU feature that is most effected by the drone being close to the wall.  

This in turn means the roll and pitch angle would also be effected since they are calculated from the gyroscope. It is predicted based on our modeling and experimental results that the drone's roll angle would be more affected from the wall being to the left or right and the drone's pitch angle would be more affected when the wall was to the front. 

For the feature generation we use a sliding window average of one second to calculate the features. We do this to mitigate general noise from the gyroscope and accelerometer. This will give us features that are more descriptive of the pose disturbance from the wall. 
  
\begin{figure}[htp]
	\centering
	\subfloat[]{\includegraphics[width=0.4\textwidth]{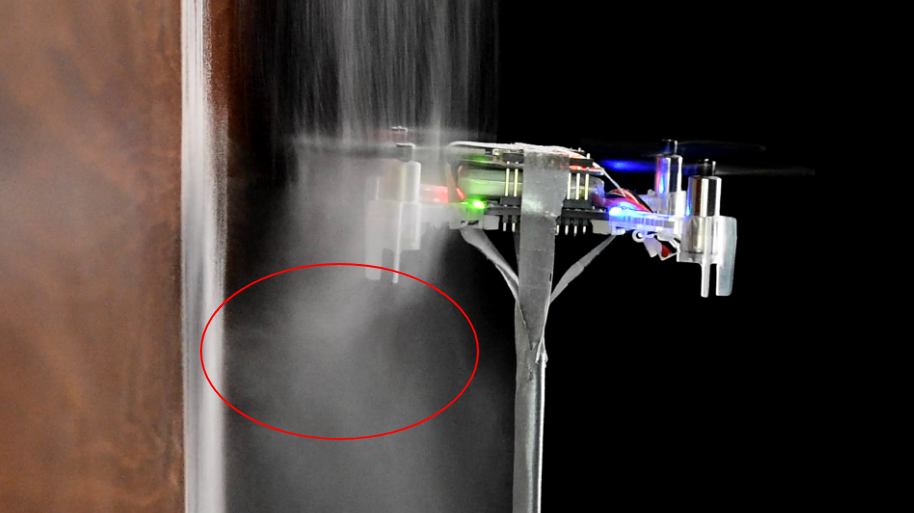}\label{fig:f2}}
	\caption{Artificial Ground Effect}
\end{figure}
 
A feature was created that calculates the drone's angle in relation to the wall. This was done as follows:

\begin{itemize}
	\item $\phi$: Roll angle in the $yz$-plane.
	\item $\psi$: Pitch angle in the $xz$-plane.
	\item $\theta$: Angle between drone and wall in the $xy$-plane.
\end{itemize}

We averaged the roll and pitch angles in a sliding window and from this a unit vector was calculated for roll and pitch as followed:
\begin{equation}
\vec{r} = \begin{bmatrix} 0 \\ \sin\phi \\ \cos\phi \end{bmatrix} \;\;\; \vec{p}=\begin{bmatrix} \sin\psi \\ 0 \\ \cos\psi \end{bmatrix}
\end{equation}
Next, the vectors $\vec{r}$ and $\vec{p}$ were added together
\begin{equation}
\vec{Z} = \vec{r} + \vec{p} = \begin{bmatrix} \sin\psi \\ \sin\phi \\ \cos\phi + \cos\psi \end{bmatrix}
\end{equation}
The vector is projected onto the $xy$-plane as that is the plane in which we are trying to measure the angle, the resulting 2D vector is
\begin{equation}
\vec{Z_p} = \begin{bmatrix} \sin\psi \\ \sin\phi \end{bmatrix}
\end{equation}
The angle theta is obtained using $\arctan$.

\begin{equation}
\theta = \arctan\left(\frac{|\sin\psi|}{|\sin\phi|}\right)
\end{equation}

As shown in Figure 5 the calculation is done with the drone centered on the $x$-axis. A result of zero would mean that the wall is directly in front and a result of $\frac{\pi}{2}$ that the wall is perpendicular to the drone. It is expected that when the wall is to the front only the pitch angle is affected and when the wall is perpendicular only the roll is affected. When we have the same reading from pitch and roll we get $\frac{\pi}{4}$ which makes sense because we would expect the same effect on roll and pitch when the drone is at such an angle.

\begin{figure}[htp]
	\centering
	\includegraphics[width=0.5\textwidth]{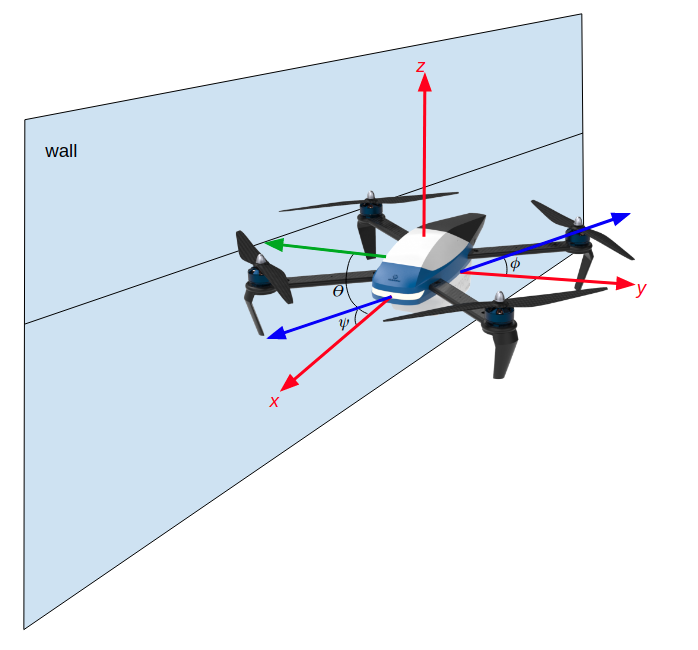}
	\caption{The drone's $x,y,z$ axes in relation to the wall}
\end{figure}

A second high-level feature was also developed. It uses the same sliding window average of the roll an pitch angles. The results from this algorithm followed more closely to the raw roll and pitch angles. The calculation is as follows:

\begin{equation}
r = \cos\phi \quad p = \cos\psi
\end{equation}
\begin{equation}
\omega = \arctan 2(p,r)
\end{equation}
We used this formulation because the cosine of the roll and pitch angles is the $\vec{k}$ component when using $\vec{i}, \vec{j}, \vec{k}$ form of the unit vectors 
employed by the previous algorithm. The $\vec{k}$ component refers the to $Z$ direction of the vector. When we take $\arctan2$ of both of these angles we get an angle that will be dependent on which angle was greater, and thus has greater effect on the drone. 
\begin{figure}[htp]
	\centering
	\subfloat[$\theta$  angle measurment ]{\includegraphics[width=0.5\textwidth]{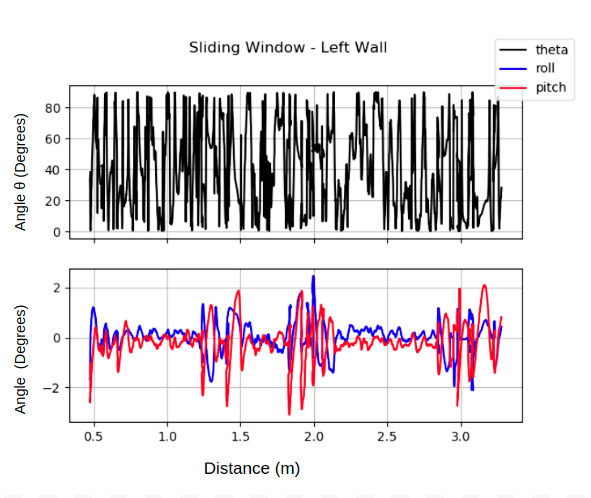}\label{fig:f1}}
	\hfill
	\subfloat[$\omega$ angle measurment ]{\includegraphics[width=0.5\textwidth]{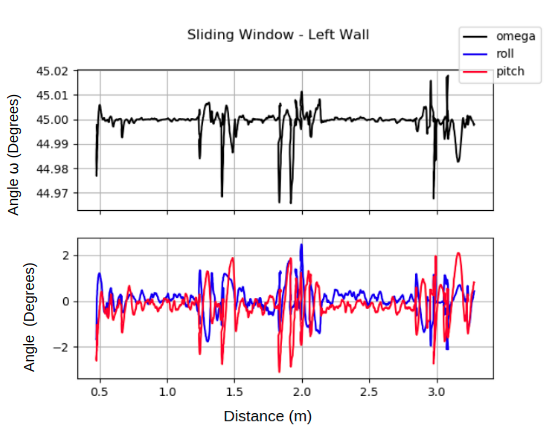}\label{fig:f2}}
	\caption{Example time sequence data for each of the two high level features. The wall was encountered to the left of the drone, at a distance within 10cm}

\end{figure}

\section{Experimentation}
\subsection{Data Collection}
Data needed to be gathered from the drone when it was flying near a wall to its left, right and front. Data was also collected from a drone flying while no walls or obstacles were near it. The drone used was a BitCraze Crazyflie 2.0 with a flow deck and loco position system (LPS) deck (fig. 2) [7]. The drone measures about 10 cm by 10 cm, and has total weight, including the decks, of 42g. The flow deck uses a laser sensor to determine the height of the quadcopter and the LPS deck triangulates with each of the LPS nodes in each corner of the flying area to determine its Cartesian position $(x,y,z)$. 

The drone was flown by a base computer using a Python interface. A driver program was created to fly the drone in a straight line in the $x$-direction when the wall was to the left, right or not present and in the $y$-direction when the wall was to the front. The wall was 1.2m long and 0.6m tall. The drone flew at a speed of 0.1 m/s and a height of 0.25m. Data was collected every hundredth of a second. We would expect to have 1200 examples but instability in the drone caused by it’s proximity to the wall, means it often flies slower than 0.1 m/s when it is along the wall, so we usually got around 1500 examples in each flight. The drone was flown along each wall side five times and flown with no wall five times for 30 seconds, producing about 3000 examples in each flight.

While the quadcopter was flying, data was collected from the onboard sensors which includes an MPU-9250 inertial measurement unit (IMU). The gyroscope sensor has digital-output based on the \(x,y,z\) axes angular rates sensors (gyroscopes) with a user-programmable full-scale range of $\pm$250, $\pm$500, $\pm$1000, and $\pm$2000$^{\circ}$/sec and integrated 16-bit ADCs. The accelerometer is a digital-output triple-axis component with a programmable full-scale range of $\pm$2g, $\pm$4g, $\pm$8g, and $\pm$16g and integrated 16-bit ADCs. In addition to the gyroscope and accelerometer data, stabilizer (roll,pitch,yaw), cartesian position $(x,y,z)$ and barometer data were collected every hundredth of a second [7].The data was stored in separate files based on the wall position or no wall present.

\subsection{Classifier Building}

The classifiers were built using the SKLearn and Pandas packages in Python [8,10]. The data was initially trimmed as follows. Based on the modeling we concluded that the gyroscope would be most affected by the drone’s proximity to the wall. As a result we dropped all features that were not the gyroscope, or stabilizer. What remained was five features: $x$-gyroscope, $y$-gyroscope, $z$-gyroscope, roll, and pitch. 

We generated 18 features to summarize the data for data mining. The data was segmented into a sliding window with one second of the time-series data. The sliding window had a 0.99 second overlap, meaning the sliding window increased by one one-hundredth of a second for each calculation of the higher level features. The 18 features were as followed:

\begin{itemize}
	\item Mean (5): Mean values of sensor, $x,y,z$ axis for gyroscope and roll and pitch angle.
	\item Standard Deviation (5): Standard deviation, $x,y,z$ gyroscope, roll and pitch angle.
	\item Mean Absolute Deviation (5): Mean absolute deviation, $x,y,z$ gyroscope, roll and pitch angle.
	\item Average Resultant (1): For each of the 100 values in the window, take the square root of the three gyroscope axis and average them together.
	\item Angle Relation to Wall (1): Average the angle for the window, calculate the unit vector of the roll and pitch angles, add them together and project the resulting vector onto the $xy$-plane. We take the $\arctan$ of the two vector components to get what angle the drone should be in relation to the wall in ideal conditions. 
	\item Cosine Angle (1): Calculate the average roll and pitch angles for the window,
	take the cosine of both averages and then take the $\arctan2$ of the resulting values. 
\end{itemize}

These first 16 features were generated because of their descriptiveness of the data from a holistic 
perspective. The final two features were generated based on the modeling presented previously in this paper.

\subsection{Experiments}

Four experiments were done using a 80\%-train 20\%-test scheme. The first experiment was simply testing wall versus no wall. The wall data used was taken from five flights of the drone flying with a wall to its left and five to its right and the no wall data was taken from five flights of the drone flying with no wall but for a longer time. The result was 16,900 examples of wall data and 17,883 examples of no wall data. 

The second experiment was testing right wall data versus left wall data. The data was taken from the same five left and right flights mentioned previously, resulting in 8,700 examples of left wall data and 8,200 examples of right wall data.

The third experiment was done using three classes which were front, right and left wall. The front wall data was taken from five flight of the drone moving in the $y$-direction (sideways) with a wall directly in front of the drone. There were the same number of left and right examples shown above and an additional 9,100 examples of front wall data. 

The final experiment used four classes which were front, right, left and no wall. Only three flights of the drone flying with no wall were used since those flights were a longer duration and the three flights were selected at random. This resulted in 8,888 examples of no wall data along with the same number of examples for the other three classes stated previously.

All four experiments evaluated the performance of three classifiers, RandomForest (RF), GradientBoosting (GB) and K-NearestNeighbors (KNN) with  $K = 5$. Each experiment was repeated 100 times on each of the three classifiers and an accuracy score was taken at each test and averaged at the end of the 100 tests. 

\section{Results}
The results of the four experiments are shown and described below. The bar graphs showing the results for each of the four experiments are shown in Figure 7.
\begin{figure*}[t]
	\centering
	\subfloat[Wall vs. No Wall]{\includegraphics[width=0.25\textwidth, height = 4cm]{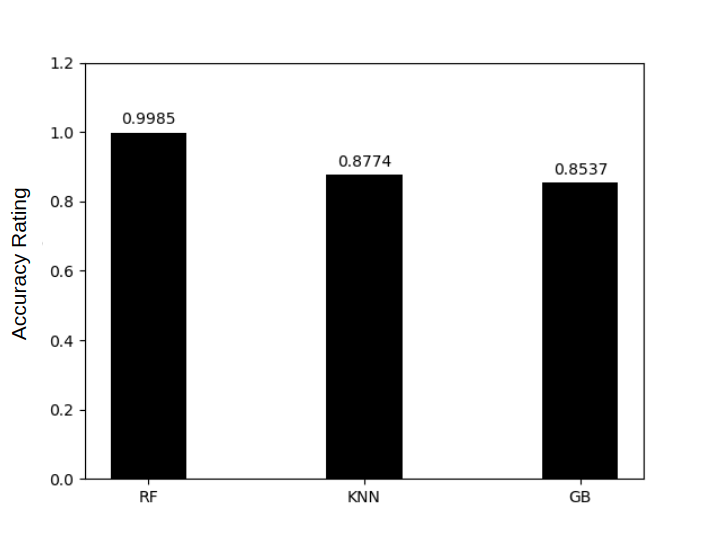}\label{fig:f1}}
	\hfill
	\subfloat[Left vs Right]{\includegraphics[width=0.25\textwidth, height = 4cm]{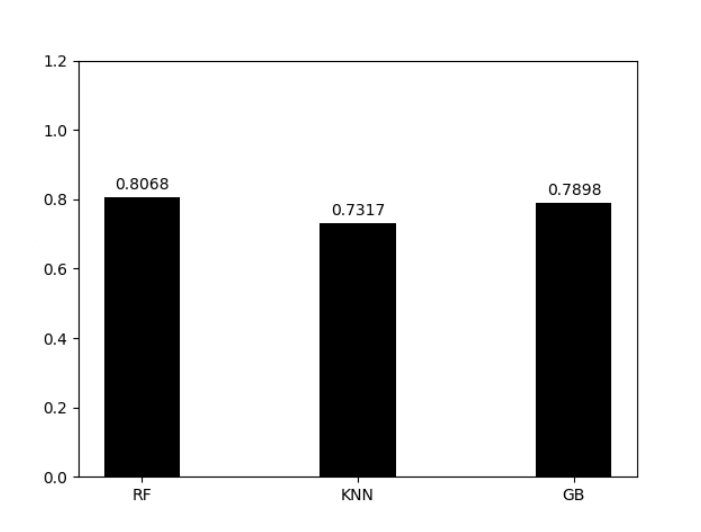}\label{fig:f2}}
	\hfill
	\subfloat[Left vs Right vs Front]{\includegraphics[width=0.25\textwidth, height = 4cm]{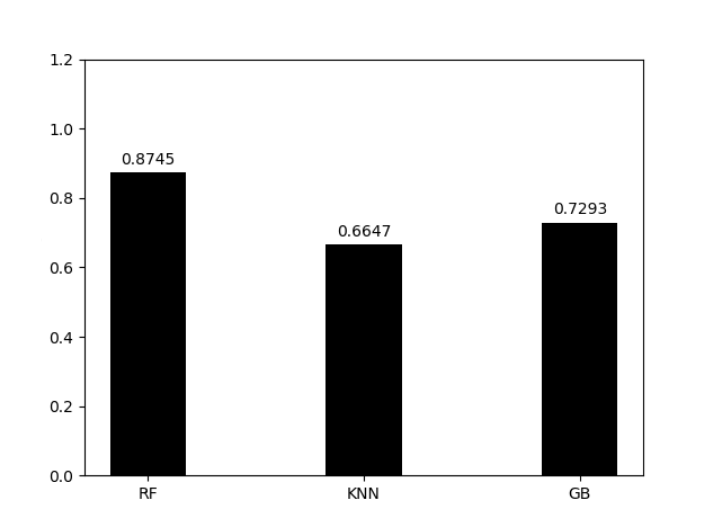}\label{fig:f3}}
	\hfill
	\subfloat[Left vs Right vs Front vs No Wall]{\includegraphics[width=0.25\textwidth, height = 4cm]{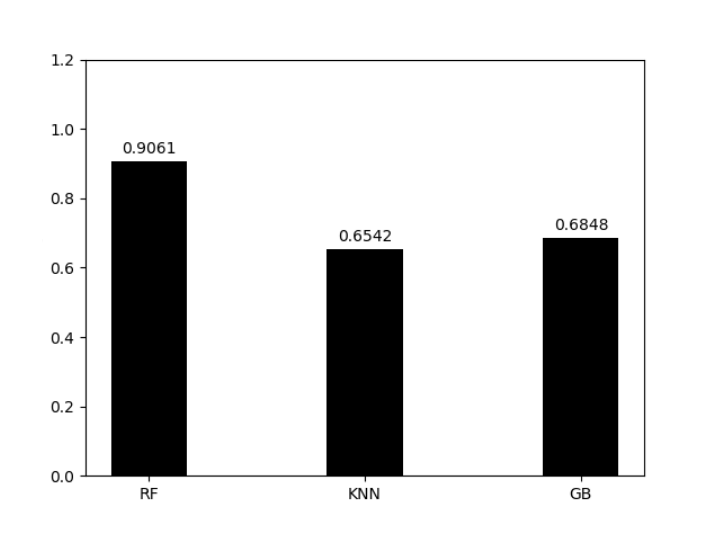}\label{fig:f4}}
	\caption{Classifier Testing Accuracies}
\end{figure*}
\subsection{Wall versus No Wall}
For the results of the first experiment the RF classifier had the highest accuracy with a 99.85\% accuracy for the average of 100 tests. The KNN classifier was second best at 87.74\% accurate followed by GB classifier at 85.34\% accurate. 

\subsection{Left versus Right Wall}
This seemed to be the hardest task for RF to classify, but it still had the highest accuracy of the three classifiers with an accuracy of 80.68\% followed by GB at 78.98\% and finally KNN at 73.17\%.

\subsection{Left versus Right versus Front}
Again the RF classifier preformed the best increasing in accuracy from the previous test to 87.45\%. It was followed by GB at 72.94\% accuracy and KNN at 66.47\% accuracy. 

\subsection{Left versus Right versus Front versus No wall}
RF continued to improve to 90.61\% accuracy while KNN and GB continued to do poorer at 65.42\% and 68.48\% accuracy respectively.

\section{Discussion}
The accuracy of the three classifiers is discussed below, as well as the implications for which would be best to implement for obstacle avoidance. We also further discuss the `artificial' ground effect that we have predicted and hypothesize is responsible for the success of our classifiers.

\subsection{Classifier Accuracy}
The RF classifier was most accurate in all 4 experiments, and the only one to preform better when testing on more than two classes when compared to left versus right testing. Contrary to this, KNN and GB both preformed worse an each consecutive experiment. All three classifiers preformed well on the first experiment. This is likely due to the clear effect the wall has on the quadcopter as shown in Figure 8.
\begin{figure}[h]
	\centering
	\includegraphics[width=0.4\textwidth, height = 5cm]{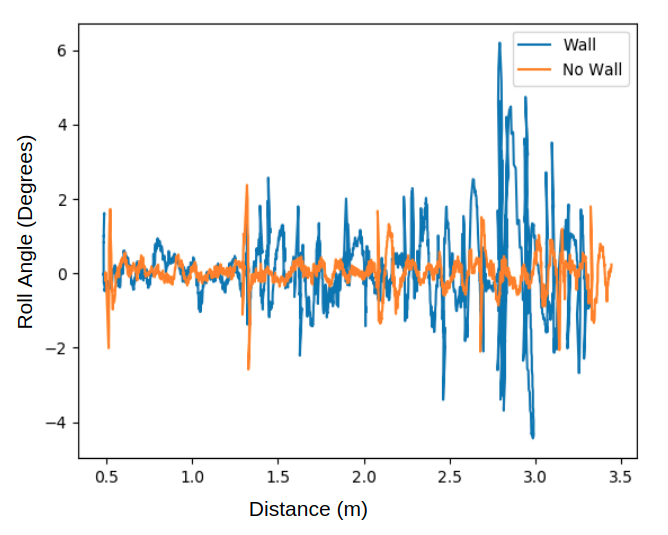}
	\caption{Wall vs. No Wall Roll Angle}
\end{figure}
When the drone is flying near a wall there is clearly visible pattern of pose disturbance as compared to flying with no wall. The left versus right wall test was the worst performer for the RF classifier. This is likely due to the fact that a wall to the right has a similar effect to that of a wall to the left. The 80\% accuracy was not bad considering the task the classifier was given. RF continued to do better in both the left, right, front experiment and the left, right, front, no wall experiment. The front wall has a different effect on the drone since the wall is on a different axis. The same applies to the final experiment with no wall because the no wall is distinctly different from when the drone is flying near a wall, whether it be to the front or side. RF's significantly better accuracy in the experiments makes is the best option to include in a driver program. While 
\textbf{it}has a long training time, it can test very quickly when the drone is in flight.



\subsection{Future Work}
Real world application could result in lower accuracy ratings. We imagine this approach to obstacle detection would be most useful in an indoor environment. Indoors the drone could encounter other wind disturbances from things like vents or moving objects in the environment. In the future we would like to include other wind disturbances in the training data when no wall is present. Seeing as the RF classifier can determine to which side a wall is, if this data is included it is possible that the classifier could also determine which pose disturbance is caused by a wall and which is not. Additionally, the classifier can be implemented into an obstacle avoidance program so that the drone can detect an obstacle using the classifier and then move around the obstacle accordingly.

\subsection{Summary}
In summary, our results show that this method can be used to accurately predict if a wall is present or not. We can also predict whether the wall is to the left right or front with decent accuracy using the RF classifier. We determined that there is more difficulty determining left wall versus right wall. Lastly, we can also predict left vs. right. vs. front. vs no wall with ~90\% accuracy. We've determined that the RandomForest classifier is best for this classification.

\end{document}